\documentclass[letterpaper,10pt,oneside, twocolumn]{article}
\RequirePackage{graphicx}
\usepackage{bbding} 
\usepackage{multirow}
\usepackage{amsfonts}
\usepackage{amsmath,bm}
\usepackage{url}
\usepackage{endnotes}
\usepackage{algorithm}
\usepackage{algorithmic}
\usepackage{booktabs}
\usepackage{sectsty}
\usepackage{cite}
\usepackage{color}
\usepackage{amssymb}
\usepackage{pifont}
\usepackage[font={small,sf},labelfont=bf]{caption}
\usepackage[hang]{subfigure}

\usepackage{balance}

\usepackage{verbatim}

\usepackage[pagebackref=false,breaklinks=true,letterpaper=true,colorlinks,bookmarks=false]{hyperref}

\newcommand{\ie}{\emph{i.e.,~}}

\newcommand{\eg}{\emph{e.g.,~}}

\usepackage{xcolor}

\usepackage{xspace}
\makeatletter
\DeclareRobustCommand\onedot{\futurelet\@let@token\@onedot}
\def\@onedot{\ifx\@let@token.\else.\null\fi\xspace}

\def\eg{\emph{e.g}\onedot} 
\def\ie{\emph{i.e}\onedot}

\allsectionsfont{\sffamily}

\def\@maketitle{\newpage
 \null
 \vskip 2em
 \begin{center}
 {\Large \sffamily  \@title \par} \vskip 1.5em {\large \lineskip .5em
\sffamily  \@name \large \lineskip .8em
\begin{tabular}[t]{c}
\sffamily  \@address
 \end{tabular}\par}
 \lineskip .8em \sffamily \textbf{}
 \end{center}
 \par
 \vskip 1.5em}

\def\title#1{\gdef\@title{#1}}
\def\name#1{\gdef\@name{#1\\}}
\def\address#1{\gdef\@address{#1}}
\gdef\@title{\uppercase{title of paper}} \gdef\@name{{\em Name of
author}\\}
\gdef\@address{Address - Line 1 \\
               Address - Line 2 \\
               Address - Line 3}

\let\@@savethanks\thanks
\def\thanks#1{\gdef\thefootnote{}\@@savethanks{#1}}
\def\sthanks#1{\gdef\thefootnote{\fnsymbol{footnote}}\@@savethanks{#1}}

\def\threeauthors#1#2#3#4{\gdef\@address{}
   \gdef\@name{#1 \\
\begin{tabular}{@{}c@{}}
        #2\relax
   \end{tabular}\hskip 0.5in
   \begin{tabular}{@{}c@{}}
        #3\relax
   \end{tabular}\hskip 0.5in
   \begin{tabular}{@{}c@{}}
        #4\relax
\end{tabular}}}

\def\twoauthors#1#2#3{\gdef\@address{}
   \gdef\@name{#1 \\
\begin{tabular}{@{}c@{}}
        #2\relax
   \end{tabular}\hskip 0.5in
   \begin{tabular}{@{}c@{}}
        #3\relax
   \end{tabular}\hskip 0.5in
}}

\title{\textbf{Renmin University of China at TRECVID 2022: Improving Video Search by Feature Fusion and Negation Understanding}}
\name{Xirong Li, Aozhu Chen, Ziyue Wang, Fan Hu, Kaibin Tian, Xinru Chen, Chengbo Dong}
\address{ MOE Key Lab of DEKE, Renmin University of China \\ AIMC Lab, School of Information, Renmin University of China \\ \url{https://ruc-aimc-lab.github.io}}
\date{}
\begin{document}
\thispagestyle{empty} \maketitle \thispagestyle{empty}

\begin{abstract}

\emph{We summarize our TRECVID 2022 Ad-hoc Video Search (AVS) experiments. Our solution is built with two new techniques, namely Lightweight Attentional Feature Fusion (LAFF) for combining diverse visual / textual features and Bidirectional Negation Learning (BNL) for addressing queries that contain negation cues. In particular, LAFF performs feature fusion at both early and late stages and at both text and video ends to exploit diverse (off-the-shelf) features. Compared to multi-head self attention, LAFF is much more compact yet more effective. Its attentional weights can also be used for selecting fewer features, with the retrieval performance mostly preserved. BNL trains a negation-aware video retrieval model by minimizing a bidirectionally constrained loss per triplet, where a triplet consists of a given training video, its original description and a partially negated description. For video feature extraction, we use pre-trained CLIP, BLIP, BEiT, ResNeXt-101 and irCSN. As for text features, we adopt bag-of-words, word2vec, CLIP and BLIP. 
Our training data consists of MSR-VTT, TGIF and VATEX that were used in our previous participation. In addition, we automatically caption the V3C1 collection for pre-training. The 2022 edition of the TRECVID benchmark has again been a fruitful participation for the RUCMM team. Our best run, with an infAP of 0.262, is ranked at the second place teamwise.}

\end{abstract}


\section{Our Approach}

Our solution for the TRECVID 2022 (TV22) AVS task is based on two newly developed techniques.
One is the Lightweight Attentional Feature Fusion (LAFF) \cite{hu2022lightweight}, an attention-based feature fusion method that performs feature fusion at both early and late stages and at both video and text ends. The other is Bidirectional Negation Learning (BNL) \cite{wang2022ntvr}, a learning based method for training a negation-aware video retrieval model, which is used to handle queries that contain negative cues, \eg ``\emph{A man is holding a knife in a non-kitchen location}''.

\subsection{LAFF-based Video Retrieval}

\begin{figure}[htbh!]
\centering
\includegraphics[width=0.6\columnwidth]{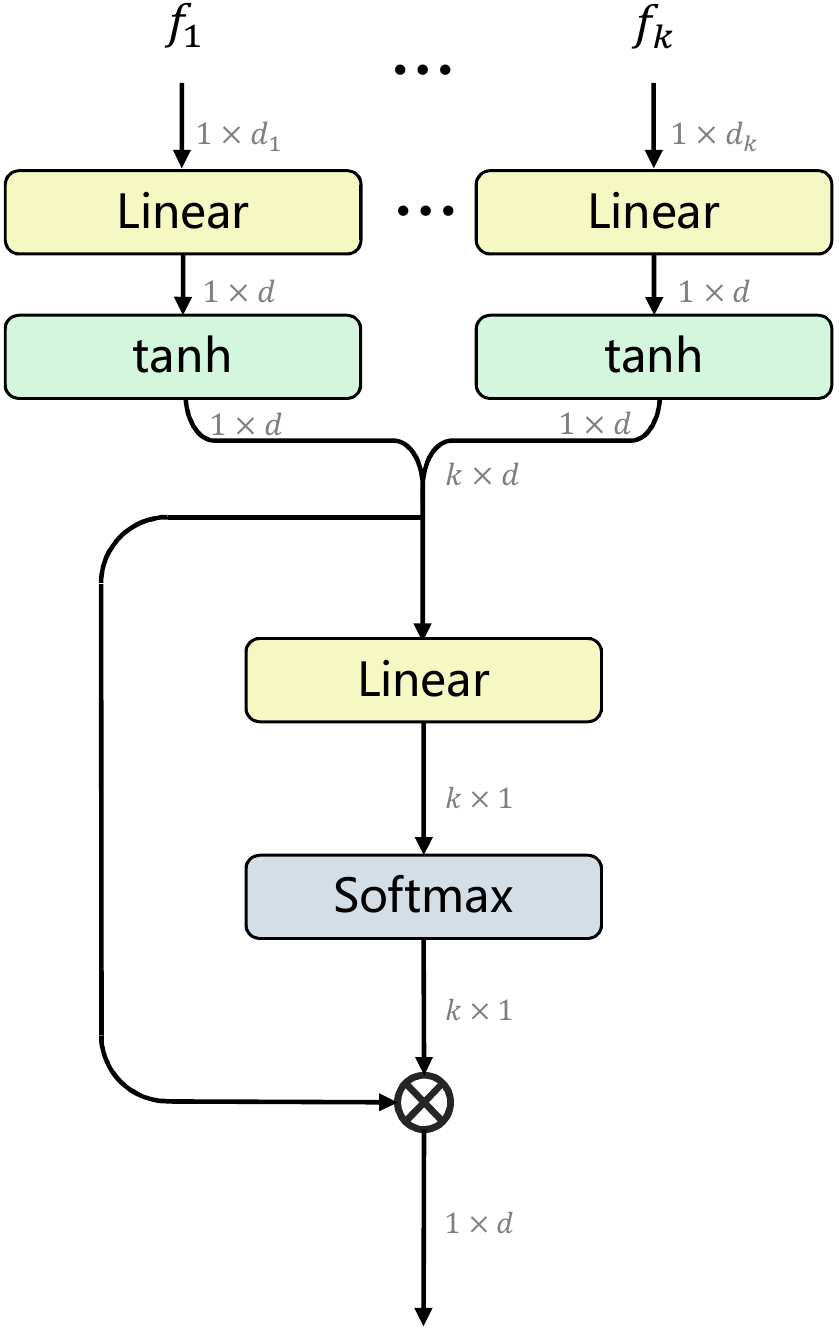}
\caption{\textbf{Lightweight Attentional Feature Fusion (LAFF)} \cite{hu2022lightweight}. }
\label{fig:framework}
\end{figure}

Given a video $x$ represented by a set of $k_1$ video-level features $\{f_{v,1},\ldots,f_{v,k_1}\}$, LAFF performs feature fusion by first  transforming each of the $k_1$ features to a $d$-dimensional feature vector and then aggregating the transformed features into a combined  feature $\bar{f}_v$ by a convex combination:
\begin{equation}
\begin{array}{ll}
     \bar{f}_v & = \mbox{LAFF}(\{f_{v,1},\ldots,f_{v,k_1}\}) \\
               & = \sum_{i=1}^{k_{1}} a_i \times Linear(f_{v,i}),
\end{array}
\end{equation}
where $Linear$ denotes a fully connected layer followed by a \textit{tanh} activation, while $\{a_i\}$ are feature-specific weights computed by a lightweight attentional mechanism, see Fig. \ref{fig:framework}. Similarly, given a textual query (or sentence) $q$ represented by a set of $k_2$ sentence-level features $\{f_{t,1},\ldots,f_{t,k_2}\}$, we obtain its combined feature $\bar{f}_t$ as $\mbox{LAFF}(\{f_{t,1},\ldots,f_{t,k_2}\}) $.

In order to compute a cross-modal similarity $s(x,q)$ between $x$ and $q$, the above video-specific and text-specific LAFFs shall be paired and trained jointly. In this work, we use $h$ pairs for computing their cross-modal sim $s(x,q)$. We use $j=1,\ldots,h$ to index each pair and the resultant features $\bar{f}_{v,j}$ and $\bar{f}_{t,j}$. Accordingly, we have 
\begin{equation}
    s(x,q) = \frac{1}{h} \sum_{j=1}^h \mbox{cosine-sim}(\bar{f}_{v,j}, \bar{f}_{t,j}).
\end{equation}



\subsubsection{Choice of Visual Features}

The following six deep visual features are used: 
\begin{enumerate}
\item  \emph{wsl}: A 2,048-d frame-level feature, extracted by ResNeXt-101 which pre-trained on weakly labeled web images followed by fine-tuning on ImageNet\footnote{\url{https://github.com/facebookresearch/WSL-Images}} \cite{wsl}. 
\item  \emph{beit}: A 1,024-d frame-level feature, extracted by BEiT which pre-trained on full ImageNet and fine-tune on 1k-class ImageNet \footnote{\url{https://github.com/microsoft/unilm/tree/master/beit}} \cite{beit}. 
\item \emph{clip}: A 768-d frame-level feature, extracted by a pre-trained CLIP (ViT-L/14)@336 model. \footnote{\url{https://github.com/openai/CLIP}} \cite{2021clip_icml}. 
\item \emph{clip-bnl}: A 512-d frame-level feature, extracted by a CLIP-\textit{bnl}, which re-trains CLIP(ViT-B/32) by bidirectional negation learning on re-purposed MSR-VTT \cite{wang2022ntvr}.
\item \emph{blip}: A 256-d frame-level feature, extracted by a BLIP(ViT-B) \cite{blip2022} which pre-trained on 129M image-text pairs \footnote{\url{https://github.com/salesforce/BLIP}}. 

\item \emph{ircsn}: A 2,048-d segment-level feature, extracted by irCSN-152~\cite{ircsn} which trained on IG-65M\footnote{\url{https://github.com/facebookresearch/VMZ/tree/master/pt}}. 
\end{enumerate}

\subsubsection{Choice of Textual Features}

We experimented with the following five sentence encoders for textual features:
\begin{enumerate}
    \item Bag-of-Words (bow)~\cite{w2vv,tmm21-sea}, which has a dimensionality of 16k according to our training data. 
    \item Word2Vec (w2v)~\cite{w2v} pretrained on Flickr tags \cite{sigir2015-hierse} .
    \item Text encoder of CLIP(ViT-L/14)@336~\cite{2021clip_icml}, which is a GPT. 
    \item Text encoder of CLIP-\textit{bnl}~\cite{wang2022ntvr}, \ie a GPT. 
    \item Text encoder of BLIP(ViT-B)~\cite{blip2022}, which is a BERT. 
\end{enumerate}


\subsubsection{Choice of (Pre-)Training Data}
Different from our TV21 system \cite{tv21-rucmm} that uses image collections 
for pre-training, this year we use a self-built video-text dataset, namely V3C1-Pseudo-Caption (V3C1-PC), as the pre-training dataset, see Table \ref{tab:V3C1-pseudo-caption}. Specifically, for each video in V3C1, we use BLIP to generate a caption for each sampled frame. A $n$-frame video will have $n$ captions. We remove duplicate captions and then use CLIP to rank the remaining captions in terms of their cross-modality similarity to the video. The top-3 ranked captions are preserved as the video's pseudo captions 

As for training data, we use the joint set of MSR-VTT~\cite{msrvtt}, TGIF~\cite{tgif} and VATEX~\cite{vatex} (M+T+V).
Following our conventional setup \cite{tv18-rucmm,tv19-rucmm,mm2019-w2vvpp,tv20-rucmm},  the development set of the TRECVID 2016 Video-to-Text Matching task~\cite{2016trecvidawad} is used as an external validation set\footnote{\url{https://github.com/li-xirong/avs}} for base model selection.

\begin{table} [htbp!]
\renewcommand{\arraystretch}{1}
\caption{\textbf{Statistics of V3C1-PC}.}

\label{tab:V3C1-pseudo-caption}
\centering
 \scalebox{1}{
 \begin{tabular}{@{} llrrrrr @{}}
\toprule

\textbf{Dataset}  &\textbf{Frames} & \textbf{Shots} &\textbf{Videos} &\textbf{Sentences}   \\
 \hline
V3C1-PC & 1,605,335 & 219,530  & 9,760 & 436,203  \\


\bottomrule
\end{tabular}
 }
\end{table}

\subsection{BNL for Negation-Aware Video Retrieval}

\begin{figure}[htbh!]
\centering
\includegraphics[width=1\columnwidth]{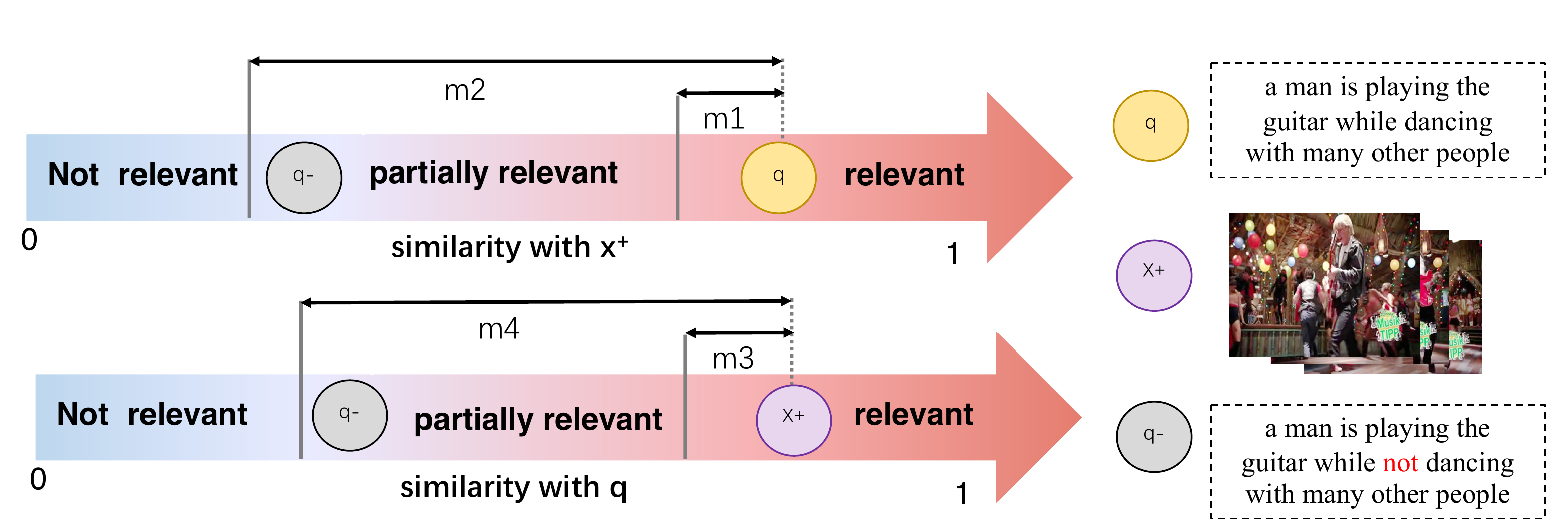}
\caption{\textbf{Bidirectional Negation Learning (BNL)}.}
\label{fig:BNl}
\end{figure}

BNL trains a negation-aware video retrieval model by letting a CLIP model learn from partially negated video descriptions. Such descriptions are automatically constructed as follows. Given a video $x^{+}$ and one of its associated original caption $q$,  a negative cue (\eg [not]) is randomly inserted into $q$, say right before an identified verb or after an auxiliary verb, to construct a partially negated video description $q^{-}$, 
as illustrated in Fig. \ref{fig:BNl}.  

Given the triplet $<x^+, q, q^->$, the following two bidirectionally constrained losses are calculated: 
\begin{equation} \label{eq:bcl1}
\begin{array}{ll}
l_{bcl}(x^+,q,q^-) = & \max (0, m_1 +s(x^+,q^-) - s(x^+, q)) + \\
    & \max (0, - m_2 - s(x^+,q^-) + s(x^+, q)),
\end{array}
\end{equation}
where $m_{1}$ and $m_{2}$ are lower and upper boundaries to bound the gap between $s(x^+,q^-)$ and $s(x^+, q)$,  with $0<m_1<m_2<2$, and 
\begin{equation} \label{eq:bcl2}
\begin{array}{ll}
l_{bcl}(q,x^+,q^-) = & \max (0, m_3 +s(q,q^-) - s(q, x^+)) + \\
    & \max (0, - m_4 - s(q, q^-) + s( q,x^+),
\end{array}
\end{equation}
where $m_{3}$ and $m_{4}$ are lower and upper boundaries to bound the gap between $s(q,q^-)$ and $s(q, x^+)$,  with $0<m_3<m_4<2$. 
The two losses are then weightedly added to a standard retrieval loss:
\begin{equation} \label{eq:bnl}
\begin{array}{ll}
l_{bnl}(q,x^+) = &\max (0, m_0 +s(x^{\#},q) - s(x^+, q))+\\
 &\lambda_1 (\ell_{bcl}(x^+,q,q^-) + \ell_{bcl}(q,x^+,q^-)), 
\end{array}
\end{equation}
where $x^{\#}$ denotes the hardest negative video and $\lambda_1$ is a small positive weight for balancing the primary and the auxiliary losses.

We use the BNL loss to  retrain CLIP(ViT-B/32) using a negation-enriched version of MSR-VTT \cite{wang2022ntvr}. The resultant model, denoted by CLIP-\emph{bnl}, is used  in the following two manners: \\
$\bullet$ As a cross-modal extractor for both video and query representation; \\
$\bullet$ As a re-ranking module specifically used for queries that have negative cues automatically detected.

\subsection{Search Result Reranking}

Given a top-ranked list of (5k) videos returned by a base model, we re-score each video in the list by considering a fine-grained cross-modal similairty between its $n$ frames and the given query. The frame-query similarity is computed based on their embeddings obtained by CLIP(ViT-L/14)@336. Max pooling is used to aggregated the frame-level similarities to the video level. The new relevance score is obtained by a weighted linear fusion of the newly computed score ($0.6$) and the original score ($0.4$). To better handle queries that have negative cues detected,  we use CLIP-\emph{bnl} instead of CLIP(ViT-L/14)@336.




\begin{figure*}[htbp!]
\centering
\includegraphics[width=1.8\columnwidth]{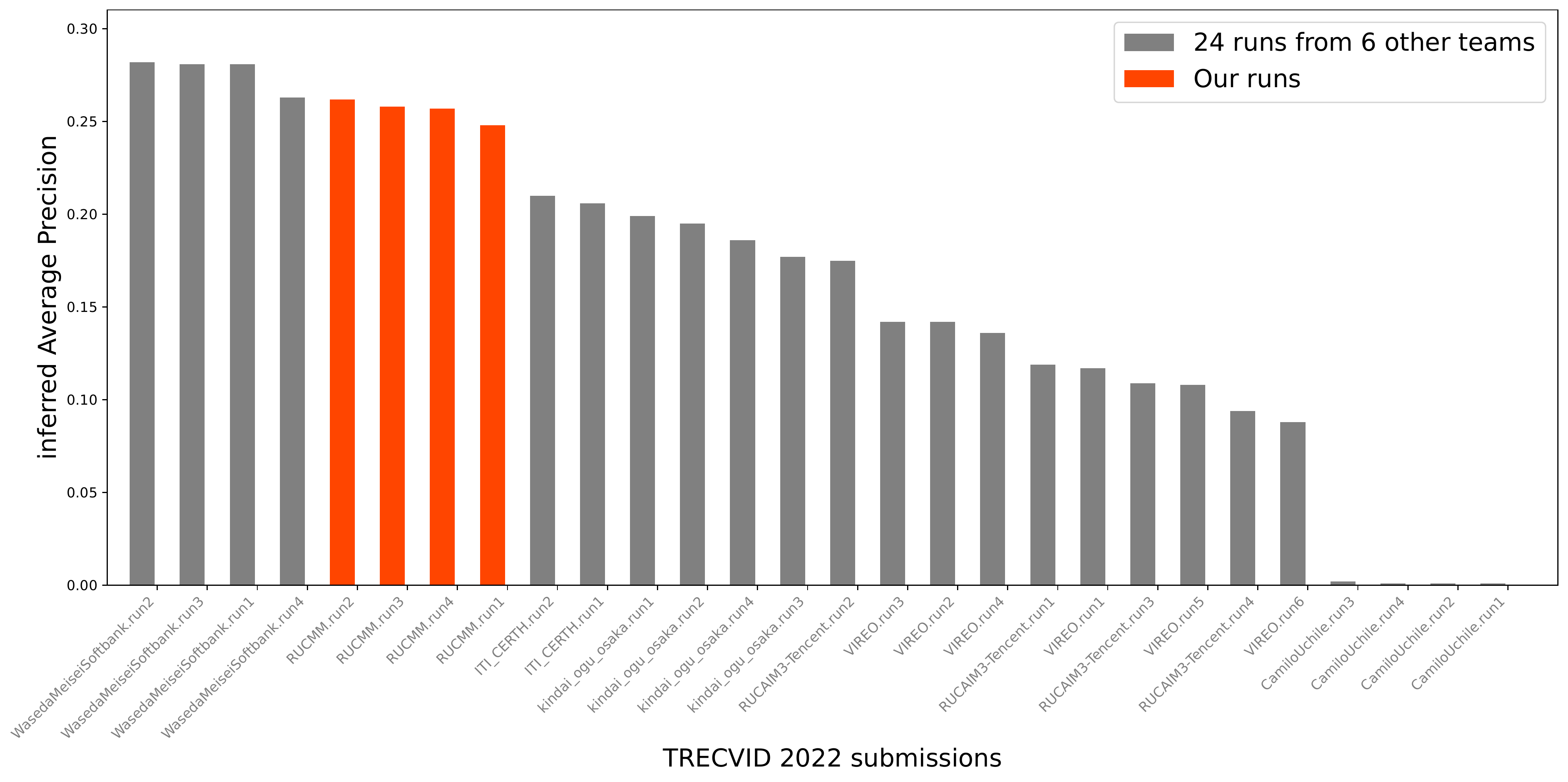}
\caption{\textbf{Overview of the TRECVID 2022 AVS benchmark evaluation}.}
\label{fig:avs-runs}
\end{figure*}

\section{Internal Evaluation}


According to our experiments on  TV16-TV18, V3C1-PC is than MS-COCO for pre-training, see Table \ref{tab:select-pretrain-data}.
\begin{table} [htbp!]
\renewcommand{\arraystretch}{1}
\caption{\textbf{Evaluating the influence of pre-training dataset on the TRECVID 16-18 AVS tasks}.}

\label{tab:select-pretrain-data}
\centering
 \scalebox{0.85}{
 \begin{tabular}{@{} llrrrrr @{}}
\toprule
\textbf{Pre-Training Dataset} & \multicolumn{1}{l}{\textbf{TV16}} & \multicolumn{1}{l}{\textbf{TV17}} & \multicolumn{1}{l}{\textbf{TV18}}  & \multicolumn{1}{l}{\textbf{MEAN}} \\
\hline
MS-COCO  & 0.263 & 0.340 & 0.174 & 0.259 \\
V3C1-PC& 0.260 & 0.356 & 0.187 & 0.268\\

\bottomrule
\end{tabular}
 }
\end{table}




To assess the influence of using CLIP-\textit{bnl} for video/text feature extraction and for search result reranking, we evaluate  the following three settings on TV16-TV21:\\
\begin{enumerate}
    \item LAFF$^{(1)}$: We pre-train LAFF on V3C1-PC and then fine-tune on M+T+V using a set of selected features\footnote{Video features: \textit{wsl}, \textit{beit},\textit{clip}, \textit{clip(ViT-B/32)}, \textit{blip} and \textit{ircsn}. Text features: \textit{bow}, \textit{w2v}, \textit{clip(ViT-B/32)},\textit{clip}, and \textit{blip}.}.
    \item LAFF$^{(2)}$: Substituting CLIP-\textit{bnl} for   CLIP (ViT-B/32) in LAFF$^{(1)}$.
    \item LAFF$^{(2)}+Re$: LAFF$^{(2)}$ + search result reranking.
\end{enumerate}


As Table \ref{tab:compare-clip-bnl-rerank} shows, LAFF$^{(2)}$ is better than LAFF$^{(1)}$, suggesting that CLIP-\emph{bnl} is beneficial. Comparing LAFF$^{(2)}$ and LAFF$^{(2)}+Re$, reranking brings in consistent improvement. 
So we rerank the search results of each base model.
\begin{table} [htbp!]
\renewcommand{\arraystretch}{1}
\caption{\textbf{Evaluating the influence of CLIP-bnl and Reranking on the TRECVID 16-21 AVS tasks}. }

\label{tab:compare-clip-bnl-rerank}
\centering
 \scalebox{0.7}{
 \begin{tabular}{@{} llrrrrrrrr @{}}
\toprule
\textbf{Model} & \multicolumn{1}{l}{\textbf{TV16}} & \multicolumn{1}{l}{\textbf{TV17}} & \multicolumn{1}{l}{\textbf{TV18}} & \multicolumn{1}{l}{\textbf{TV19}} & \multicolumn{1}{l}{\textbf{TV20}} & \multicolumn{1}{l}{\textbf{TV21}} & \multicolumn{1}{l}{\textbf{MEAN}} \\
\hline
LAFF$^{(1)}$  & 0.259 & 0.35 & 0.189 & 0.238 & 0.359 & 0.353 & 0.291 \\
LAFF$^{(2)}$ & 0.262 & 0.357 & 0.193 & 0.243 & 0.357 & 0.363 & 0.296 \\
LAFF$^{(2)}$+Re & 0.276 & 0.361 & 0.192 & 0.255 & 0.361 & 0.365 & 0.302\\
\bottomrule
\end{tabular}
 }
\end{table}






\section{Submissions}

We submit the best performance setting\footnote{Video features: \textit{wsl}, \textit{beit},\textit{clip}, \textit{clip-bnl}, \textit{blip} and \textit{ircsn}. Text features: \textit{bow}, \textit{w2v}, ,\textit{clip},\textit{clip-bnl} and \textit{blip}. Pre-training data: V3C1-PC. Training data: M+T+V} on the TV16-TV21 as Run 4. As for Run 3, we remove two heavy text features (\ie bow and w2v) to see whether those heavy text encoders could be removed. As for Run 2, we fuse the results of narrative of queries by LAFF (w/o bow and w2v) and Run3. As for Run 1, we tried a naive query augmentation strategy by automatically appending noun / adjective based keywords at the end of each query.
Our four runs are as follows:
\begin{itemize}
    \item \emph{Run 4}: LAFF 
    \item \emph{Run 3}: LAFF (w/o bow and w2v)
    \item \emph{Run 2}: Late average fusion of Run3 on test queries and narrative of queries.
    \item \emph{Run 1}: Late average fusion of multiple augmented query retrieval results.
\end{itemize}

The performance of our four runs on the TV22 AVS task is shown in Fig. \ref{fig:avs-runs}. Our best run is Run 2, which with a mean infAP of 0.262, is ranked second teamwise.
\section*{Acknowledgments}
The authors are grateful to the TRECVID coordinators for the benchmark organization effort \cite{tv22}. This research was supported by the National Natural Science Foundation of China (No. 62172420, No. 61672523), Beijing Natural Science Foundation (No. 4202033), and the Fundamental Research Funds for the Central Universities and the Research Funds of Renmin University of China (No. 18XNLG19).

\bibliographystyle{latex8}
\balance
\bibliography{tv22}

\begin{thebibliography}{10}\setlength{\itemsep}{-1ex}\small

\bibitem{tv22}
G.~Awad.
\newblock {TRECVID} 2022.
\newblock \url{https://www-nlpir.nist.gov/projects/tv2022/}, 2022.

\bibitem{2016trecvidawad}
G.~Awad, J.~Fiscus, D.~Joy, M.~Michel, A.~Smeaton, W.~Kraaij, G.~Qu\'{e}not,
  M.~Eskevich, R.~Aly, R.~Ordelman, G.~Jones, B.~Huet, and M.~Larson.
\newblock Trecvid 2016: Evaluating video search, video event detection,
  localization, and hyperlinking.
\newblock In {\em TRECVID Workshop}, 2016.

\bibitem{beit}
H.~Bao, L.~Dong, and F.~Wei.
\newblock {BEiT}: {BERT} pre-training of image transformers.
\newblock 2021.

\bibitem{w2vv}
J.~Dong, X.~Li, and C.~G.~M. Snoek.
\newblock Predicting visual features from text for image and video caption
  retrieval.
\newblock {\em IEEE Transactions on Multimedia}, 2018.

\bibitem{ircsn}
D.~Ghadiyaram, D.~Tran, and M.~Feiszli.
\newblock Large-scale weakly-supervised pre-training for video action
  recognition.
\newblock In {\em CVPR}, 2019.

\bibitem{hu2022lightweight}
F.~Hu, A.~Chen, Z.~Wang, F.~Zhou, J.~Dong, and X.~Li.
\newblock Lightweight attentional feature fusion: A new baseline for
  text-to-video retrieval.
\newblock In {\em ECCV}, 2022.

\bibitem{blip2022}
J.~Li, D.~Li, C.~Xiong, and S.~C.~H. Hoi.
\newblock {BLIP:} bootstrapping language-image pre-training for unified
  vision-language understanding and generation.
\newblock In {\em ICML}, 2022.

\bibitem{tv21-rucmm}
X.~Li, A.~Chen, F.~Hu, X.~Chen, C.~Dong, and G.~Yang.
\newblock Renmin {U}niversity of china at trecvid 2021: Searching and
  describing video.
\newblock In {\em TRECVID Workshop}, 2021.

\bibitem{tv18-rucmm}
X.~Li, J.~Dong, C.~Xu, J.~Cao, X.~Wang, and G.~Yang.
\newblock Renmin {U}niversity of {C}hina and {Z}hejiang {G}ongshang
  {U}niversity at {TRECVID} 2018: Deep cross-modal embeddings for video-text
  retrieval.
\newblock In {\em TRECVID Workshop}, 2018.

\bibitem{sigir2015-hierse}
X.~Li, S.~Liao, W.~Lan, X.~Du, and G.~Yang.
\newblock Zero-shot image tagging by hierarchical semantic embedding.
\newblock In {\em SIGIR}, 2015.

\bibitem{mm2019-w2vvpp}
X.~Li, C.~Xu, G.~Yang, Z.~Chen, and J.~Dong.
\newblock {W2VV}++: Fully deep learning for ad-hoc video search.
\newblock In {\em ACMMM}, 2019.

\bibitem{tv19-rucmm}
X.~Li, J.~Ye, C.~Xu, S.~Yun, L.~Zhang, X.~Wang, R.~Qian, and J.~Dong.
\newblock Renmin {U}niversity of china and {Z}hejiang {G}ongshang {U}niversity
  at {TRECVID} 2019: Learn to search and describe videos.
\newblock In {\em TRECVID Workshop}, 2019.

\bibitem{tv20-rucmm}
X.~Li, F.~Zhou, and A.~Chen.
\newblock {Renmin} {U}niversity of {C}hina at {TRECVID} 2020: Sentence encoder
  assembly for ad-hoc video search.
\newblock In {\em TRECVID Workshop}, 2020.

\bibitem{tmm21-sea}
X.~Li, F.~Zhou, C.~Xu, J.~Ji, and G.~Yang.
\newblock {SEA}: Sentence encoder assembly for video retrieval by textual
  queries.
\newblock {\em IEEE Transactions on Multimedia}, 23:4351--4362, 2021.

\bibitem{tgif}
Y.~Li, Y.~Song, L.~Cao, J.~Tetreault, L.~Goldberg, A.~Jaimes, and J.~Luo.
\newblock {TGIF}: A new dataset and benchmark on animated {GIF} description.
\newblock In {\em CVPR}, 2016.

\bibitem{wsl}
D.~Mahajan, R.~Girshick, V.~Ramanathan, K.~He, M.~Paluri, Y.~Li, Y.~Bharambe,
  and L.~Maaten.
\newblock Exploring the limits of weakly supervised pretraining.
\newblock In {\em ECCV}, 2018.

\bibitem{w2v}
T.~Mikolov, K.~Chen, G.~Corrado, and J.~Dean.
\newblock Efficient estimation of word representations in vector space.
\newblock In {\em ICLR}, 2013.

\bibitem{2021clip_icml}
A.~Radford, J.~W. Kim, C.~Hallacy, A.~Ramesh, G.~Goh, S.~Agarwal, G.~Sastry,
  A.~Askell, P.~Mishkin, J.~Clark, G.~Krueger, and I.~Sutskever.
\newblock Learning transferable visual models from natural language
  supervision.
\newblock In {\em ICML}, 2021.

\bibitem{vatex}
X.~Wang, J.~Wu, J.~Chen, L.~Li, Y.~F. Wang, and W.~Y. Wang.
\newblock Vatex: A large-scale, high-quality multilingual dataset for
  video-and-language research.
\newblock In {\em ICCV}, 2019.

\bibitem{wang2022ntvr}
Z.~Wang, A.~Chen, F.~Hu, and X.~Li.
\newblock Learn to understand negation in video retrieval.
\newblock In {\em ACMMMM}, 2022.

\bibitem{msrvtt}
J.~Xu, T.~Mei, T.~Yao, and Y.~Rui.
\newblock {MSR-VTT}: {A} large video description dataset for bridging video and
  language.
\newblock In {\em CVPR}, 2016.

\end{thebibliography}

\end{document}